%% file: reptile.tex
\documentclass[11pt]{article} 
\usepackage[american]{babel}
\usepackage[dvipsnames]{xcolor}
\usepackage[left=1in,right=1in,top=1in,bottom=1in]{geometry}
\usepackage{hyperref}
 \hypersetup{
   colorlinks   = true, 
   urlcolor     = blue, 
   linkcolor    = blue, 
   citecolor   = black 
 }
\usepackage{url}
\usepackage{amsmath,amssymb,mathtools,amsthm}
\usepackage{bm}
\usepackage{mdframed}
\usepackage{algpseudocode}
\usepackage{algorithm}
\usepackage[font={small}]{caption}
\usepackage{subcaption}

\input{rl_math}

\input{bold_symbols}
\input{exercises_def}

\usepackage{cleveref}

\title{On First-Order Meta-Learning Algorithms}
\newcommand{\dbyd}[1]{\frac{\partial}{\partial#1}}
\newcommand{\gmaml}{g_{\mathrm{MAML}}}
\newcommand{\grep}{g_{\mathrm{Reptile}}}
\newcommand{\gfom}{g_{\mathrm{FOMAML}}}
\newcommand{\cW}{\mathcal{W}}

 \newcommand{\ft}{f_{\tau}}

\newcommand{\gradph}{\grad_{\phi}}
\newcommand{\lt}{L_{\tau}}
\renewcommand{\half}{\tfrac{1}{2}}
\newcommand{\first}{$1^{\text{st}}$}
\newcommand{\ith}{i\textsuperscript{th}}

\makeatletter
\newcommand*\wt[1]{\mathpalette\wthelper{#1}}
\newcommand*\wthelper[2]{%
        \hbox{\dimen@\accentfontxheight#1%
                \accentfontxheight#11.3\dimen@
                $\m@th#1\widetilde{#2}$%
                \accentfontxheight#1\dimen@
        }%
}

\newcommand*\accentfontxheight[1]{%
        \fontdimen5\ifx#1\displaystyle
                \textfont
        \else\ifx#1\textstyle
                \textfont
        \else\ifx#1\scriptstyle
                \scriptfont
        \else
                \scriptscriptfont
        \fi\fi\fi3
}
\makeatother

\newcommand{\tilphi}{\wt{\phi}}

\author{Alex Nichol and Joshua Achiam and John Schulman \\
OpenAI\\
\texttt{\{alex, jachiam, joschu\}@openai.com}
\date{}
}

\begin{document}

\maketitle

\begin{abstract}
This paper considers meta-learning problems, where there is a distribution of tasks, and we would like to obtain an agent that performs well (i.e., learns quickly) when presented with a previously unseen task sampled from this distribution. We analyze a family of algorithms for learning a parameter initialization that can be fine-tuned quickly on a new task, using only first-order derivatives for the meta-learning updates. This family includes and generalizes first-order MAML, an approximation to MAML obtained by ignoring second-order derivatives. It also includes Reptile, a new algorithm that we introduce here, which works by repeatedly sampling a task, training on it, and moving the initialization towards the trained weights on that task. We expand on the results from Finn et al. showing that first-order meta-learning algorithms perform well on some well-established benchmarks for few-shot classification, and we provide theoretical analysis aimed at understanding why these algorithms work.
\end{abstract}

\section{Introduction}

While machine learning systems have surpassed humans at many tasks, they generally need far more data to reach the same level of performance.
For example, Schmidt et al. \cite{schmidt2009meaning,salakhutdinov2012one} showed that human subjects can recognize new object categories based on a few example images.
Lake et al. \cite{lake2015human} noted that on the Atari game of Frostbite, human novices were able to make significant progress on the game after 15 minutes, but double-dueling-DQN \cite{wang2015dueling} required more than 1000 times more experience to attain the same score.

It is not completely fair to compare humans to algorithms learning from scratch, since humans enter the task with a large amount of prior knowledge, encoded in their brains and DNA. Rather than learning from scratch, they are fine-tuning and recombining a set of pre-existing skills.
The work cited above, by Tenenbaum and collaborators, argues that humans' fast-learning abilities can be explained as Bayesian inference, and that the key to developing algorithms with human-level learning speed is to make our algorithms more Bayesian.
However, in practice, it is challenging to develop (from first principles) Bayesian machine learning algorithms that make use of deep neural networks and are computationally feasible.

Meta-learning has emerged recently as an approach for learning from small amounts of data.
Rather than trying to emulate Bayesian inference (which may be computationally intractable), meta-learning seeks to directly optimize a fast-learning algorithm, using a dataset of tasks.
Specifically, we assume access to a distribution over tasks, where each task is, for example, a classification problem. 
From this distribution, we sample a training set and a test set of tasks. 
Our algorithm is fed the training set, and it must produce an agent that has good average performance on the test set.
Since each task corresponds to a learning problem, performing well on a task corresponds to learning quickly.

A variety of different approaches to meta-learning have been proposed, each with its own pros and cons.
In one approach, the learning algorithm is encoded in the weights of a recurrent network, but gradient descent is not performed at test time.
This approach was proposed by Hochreiter et al. \cite{hochreiter2001learning} who used LSTMs for next-step prediction and has been followed up by a burst of recent work, for example, Santoro et al. \cite{santoro2016meta} on few-shot classification, and Duan et al. \cite{duan2016rl} for the POMDP setting.

A second approach is to learn the initialization of a network, which is then fine-tuned at test time on the new task. A classic example of this approach is pretraining using a large dataset (such as ImageNet \cite{deng2009imagenet}) and fine-tuning on a smaller dataset (such as a dataset of different species of bird \cite{zhang2014part}).
However, this classic pre-training approach has no guarantee of learning an initialization that is good for fine-tuning, and ad-hoc tricks are required for good performance.
More recently, Finn et al. \cite{finn2017model} proposed an algorithm called MAML, which directly optimizes performance with respect to this initialization---differentiating through the fine-tuning process.
In this approach, the learner falls back on a sensible gradient-based learning algorithm even when it receives out-of-sample data, thus allowing it to generalize better than the RNN-based approaches \cite{finn2017meta}.
On the other hand, since MAML needs to differentiate through the optimization process, it's not a good match for problems where we need to perform a large number of gradient steps at test time.
The authors also proposed a variant called first-order MAML (FOMAML), which is defined by ignoring the second derivative terms, avoiding this problem but at the expense of losing some gradient information. Surprisingly, though, they found that FOMAML worked nearly as well as MAML on the Mini-ImageNet dataset \cite{vinyals2016matching}.
(This result was foreshadowed by prior work in meta-learning \cite{andrychowicz2016learning,ravi2017optimization} that ignored second derivatives when differentiating through gradient descent, without ill effect.)
In this work, we expand on that insight and explore the potential of meta-learning algorithms based on first-order gradient information, motivated by the potential applicability to problems where it's too cumbersome to apply techniques that rely on higher-order gradients (like full MAML). 

We make the following contributions:
\begin{itemize}
\item We point out that first-order MAML \cite{finn2017model} is simpler to implement than was widely recognized prior to this article.
\item We introduce Reptile, an algorithm closely related to FOMAML, which is equally simple to implement. 
Reptile is so similar to joint training (i.e., training to minimize loss on the expecation over training tasks) that it is especially surprising that it works as a meta-learning algorithm.
Unlike FOMAML, Reptile doesn't need a training-test split for each task, which may make it a more natural choice in certain settings.
It is also related to the older idea of fast weights / slow weights \cite{hinton1987using}.
\item We provide a theoretical analysis that applies to both first-order MAML and Reptile, showing that they both optimize for within-task generalization.
\item On the basis of empirical evaluation on the Mini-ImageNet \cite{vinyals2016matching} and Omniglot \cite{lake2011one} datasets, we provide some insights for best practices in implementation.
\end{itemize}

\section{Meta-Learning an Initialization} \label{mlai}

We consider the optimization problem of MAML \cite{finn2017model}: find an initial set of parameters, $\phi$, such that for a randomly sampled task $\tau$ with corresponding loss $L_{\tau}$, the learner will have low loss after $k$ updates. That is:
\begin{align}
\minimize_{\phi} \Eb{\tau}{L_{\tau}\left(U_{\tau}^k (\phi) \right)},
\label{metalearn}
\end{align}
where $U^k_{\tau}$ is the operator that updates $\phi$ $k$ times using data sampled from $\tau$.
In few-shot learning, $U$ corresponds to performing gradient descent or Adam \cite{kingma2015adam} on batches of data sampled from $\tau$.

MAML solves a version of \Cref{metalearn} that makes on additional assumption: for a given task $\tau$, the inner-loop optimization uses training samples $A$, whereas the loss is computed using test samples $B$. This way, MAML optimizes for generalization, akin to cross-validation. Omitting the superscript $k$, we notate this as
\begin{align}
\minimize_{\phi} \Eb{\tau}{L_{\tau, B}\left(U_{\tau, A} (\phi) \right)},
\label{maml_opt}
\end{align}
MAML works by optimizing this loss through stochastic gradient descent, i.e., computing
\begin{align}
\gmaml &= \dbyd{\phi} L_{\tau, B}\lrparen*{U_{\tau, A} (\phi)} \\
&= U'_{\tau, A}(\phi) L'_{\tau, B}(\tilphi), \quad\text{where}\quad \tilphi = U_{\tau, A} (\phi) \label{mamlchainrule}
\end{align}
In \Cref{mamlchainrule}, $U'_{\tau, A}(\phi)$ is the Jacobian matrix of the update operation $U_{\tau, A}$.
$U_{\tau, A}$ corresponds to adding a sequence of gradient vectors to the initial vector, i.e., $U_{\tau, A}(\phi) = \phi + g_1 + g_2 + \dots + g_{k}$. (In Adam, the gradients are also rescaled elementwise, but that does not change the conclusions.)
First-order MAML (FOMAML) treats these gradients as constants, thus, it replaces Jacobian $U'_{\tau, A}(\phi)$ by the identity operation.
Hence, the gradient used by FOMAML in the outer-loop optimization is $\gfom = L'_{\tau, B}(\tilphi)$.
Therefore, FOMAML can be implemented in a particularly simple way: (1) sample task $\tau$; (2) apply the update operator, yielding $\tilphi = U_{\tau, A}(\phi)$; (3) compute the gradient at $\tilphi$, $\gfom = L'_{\tau, B}(\tilphi)$; and finally (4) plug $\gfom$ into the outer-loop optimizer.

\section{Reptile}

In this section, we describe a new first-order gradient-based meta-learning algorithm called Reptile.
Like MAML, Reptile learns an initialization for the parameters of a neural network model, such that when we optimize these parameters at test time, learning is fast---i.e., the model generalizes from a small number of examples from the test task.
The Reptile algorithm is as follows:
\newcommand{\avgl}{\overline{L}}
\begin{algorithm}[H]
\caption{Reptile (serial version)}
\begin{algorithmic}
\State Initialize $\phi$, the vector of initial parameters
\For{iteration = $1,2,\dots$}
  \State Sample task $\tau$, corresponding to loss $L_{\tau}$ on weight vectors $\tilphi$
  \State Compute $\tilphi = U^k_{\tau}(\phi)$, denoting $k$ steps of SGD or Adam
  \State Update $\phi \leftarrow \phi + \epsilon (\tilphi - \phi)$
\EndFor
\end{algorithmic}
\end{algorithm}
In the last step, instead of simply updating $\phi$ in the direction $\tilphi - \phi$, we can treat $(\phi - \tilphi)$ as a gradient and plug it into an adaptive algorithm such as Adam \cite{kingma2015adam}.
(Actually, as we will discuss in \Cref{sec:updateexp}, it is most natural to define the Reptile gradient as $(\phi -\tilphi)/\alpha$, where $\alpha$ is the stepsize used by the SGD operation.)
We can also define a parallel or batch version of the algorithm that evaluates on $n$ tasks each iteration and updates the initialization to
\begin{align}
\phi \leftarrow \phi + \epsilon \frac{1}{n}\sum_{i=1}^n(\tilphi_i - \phi)
\end{align}
where $\tilphi_i = U^k_{\tau_i}(\phi)$; the updated parameters on the $\ith$ task.

This algorithm looks remarkably similar to joint training on the expected loss $\Eb{\tau}{\lt}$.
Indeed, if we define $U$ to be a single step of gradient descent ($k=1$), then this algorithm corresponds to stochastic gradient descent on the expected loss:
\begin{align}
  g_{\text{Reptile},k=1} &= \Eb{\tau}{\phi - U_{\tau}(\phi)} / \alpha \\
  &= \Eb{\tau}{\gradph L_{\tau}(\phi)} 
\end{align}
However, if we perform multiple gradient updates in the partial minimization $(k >1)$, then the expected update $\Eb{\tau}{U^k_{\tau}(\phi)}$ does not correspond to taking a gradient step on the expected loss $\Eb{\tau}{L_{\tau}}$.
Instead, the update includes important terms coming from second-and-higher derivatives of $L_{\tau}$, as we will analyze in \Cref{sec:updateexp}.
 Hence, Reptile converges to a solution that's very different from the minimizer of the expected loss $\Eb{\tau}{\lt}$.

Other than the stepsize parameter $\epsilon$ and task sampling, the batched version of Reptile is the same as the SimuParallelSGD algorithm \cite{zinkevich2010parallelized}. SimuParallelSGD is a method for communication-efficient distributed optimization, where workers perform gradient updates locally and infrequently average their parameters, rather than the standard approach of averaging gradients.

\section{Case Study: One-Dimensional Sine Wave Regression}
\newcommand{\xonetop}{(x_1, y_1), (x_2, y_2), \dots, (x_p, y_p)}

As a simple case study, let's consider the 1D sine wave regression problem, which is slightly modified from Finn et al. \cite{finn2017model}.
This problem is instructive since by design, joint training can't learn a very useful initialization; however, meta-learning methods can.
\begin{itemize}
  \item The task $\tau=(a, b)$ is defined by the amplitude $a$ and phase $\phi$ of a sine wave function $\ft(x) = a \sin(x + b)$. The task distribution by sampling $a \sim U([0.1, 5.0])$ and $b \sim U([0, 2\pi])$.
  \item Sample $p$ points $x_1, x_2, \dots, x_p \sim U([-5, 5])$
  \item Learner sees $\xonetop$ and predicts the whole function $f(x)$
  \item Loss is $\ell_2$ error on the whole interval $[-5, 5]$
  \begin{align}
    \lt(f) = \int_{-5}^{5}\mathrm{d}x \norm*{f(x) - \ft(x)}^2
  \end{align}
  We calculate this integral using $50$ equally-spaced points $x$.
\end{itemize}
First note that the average function is zero everywhere, i.e., $\Eb{\tau}{\ft(x)}=0$, due to the random phase $b$.
Therefore, it is useless to train on the expected loss $\Eb{\tau}{\lt}$, as this loss is minimized by the zero function $f(x)=0$.

On the other hand, MAML and Reptile give us an initialization that outputs approximately $f(x)=0$ before training on a task $\tau$, but the internal feature representations of the network are such that after training on the sampled datapoints $\xonetop$, it closely approximates the target function $f_{\tau}$.
This learning progress is shown in the figures below.
\Cref{fig:fastlearnsine} shows that after Reptile training, the network can quickly converge to a sampled sine wave and infer the values away from the sampled points. As points of comparison, we also show the behaviors of MAML and a randomly-initialized network on the same task.
\begin{figure}[!h]
\centering
\begin{subfigure}[b]{.3\textwidth}
\includegraphics[width=\textwidth]{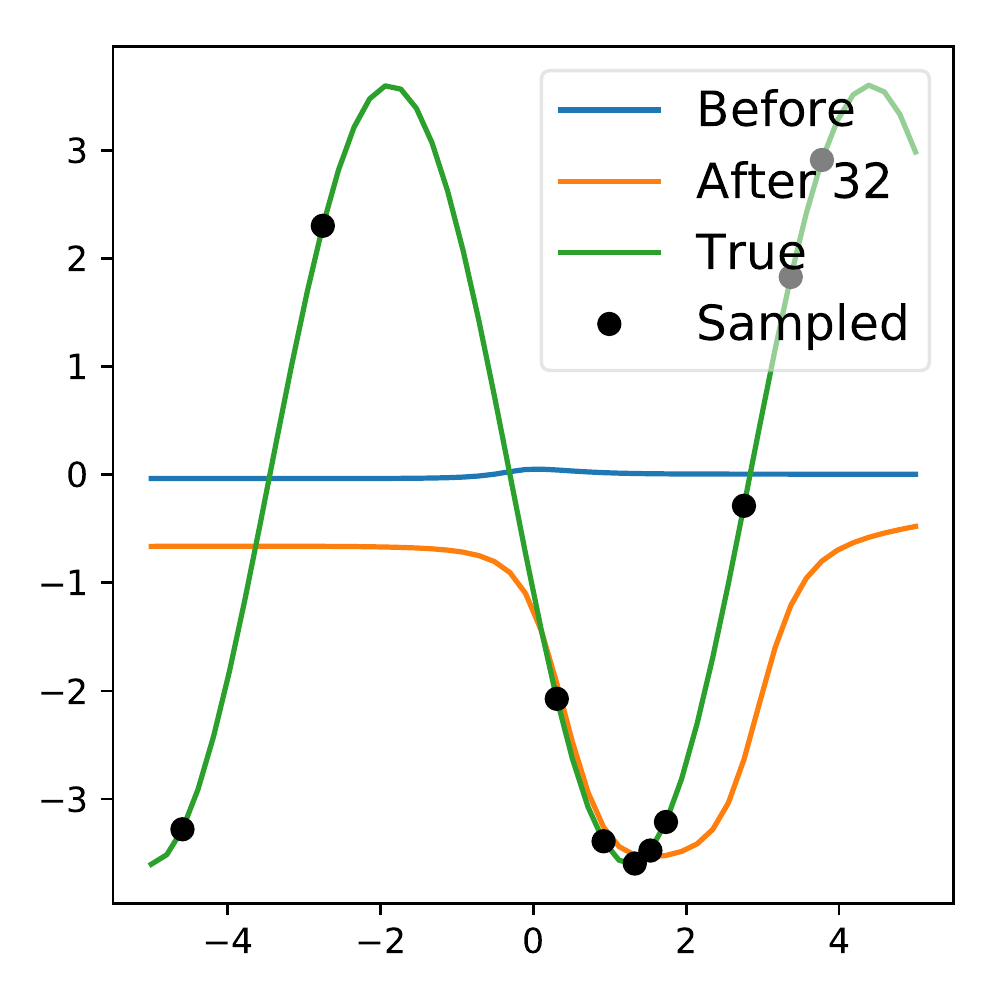}
\caption{Before training}
\end{subfigure}
\begin{subfigure}[b]{.3\textwidth}
\includegraphics[width=\textwidth]{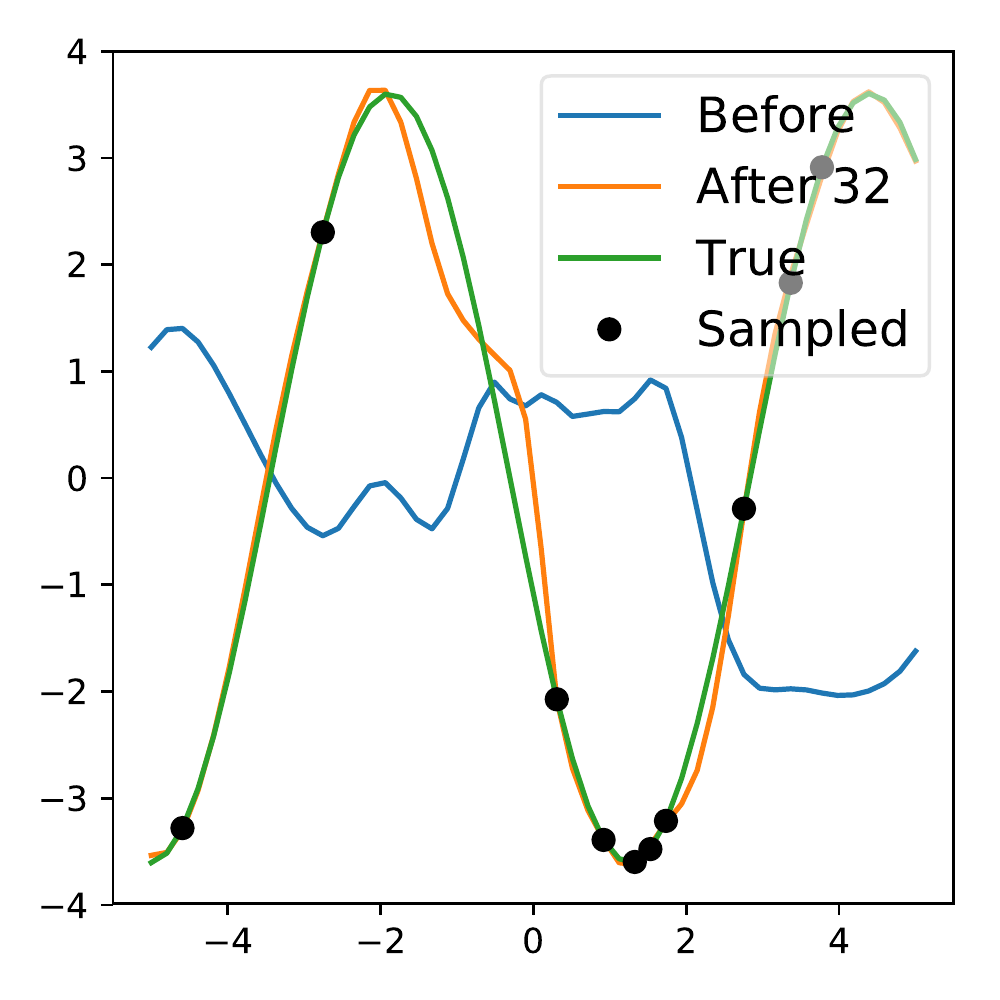}
\caption{After MAML training}
\end{subfigure}
\begin{subfigure}[b]{.3\textwidth}
\includegraphics[width=\textwidth]{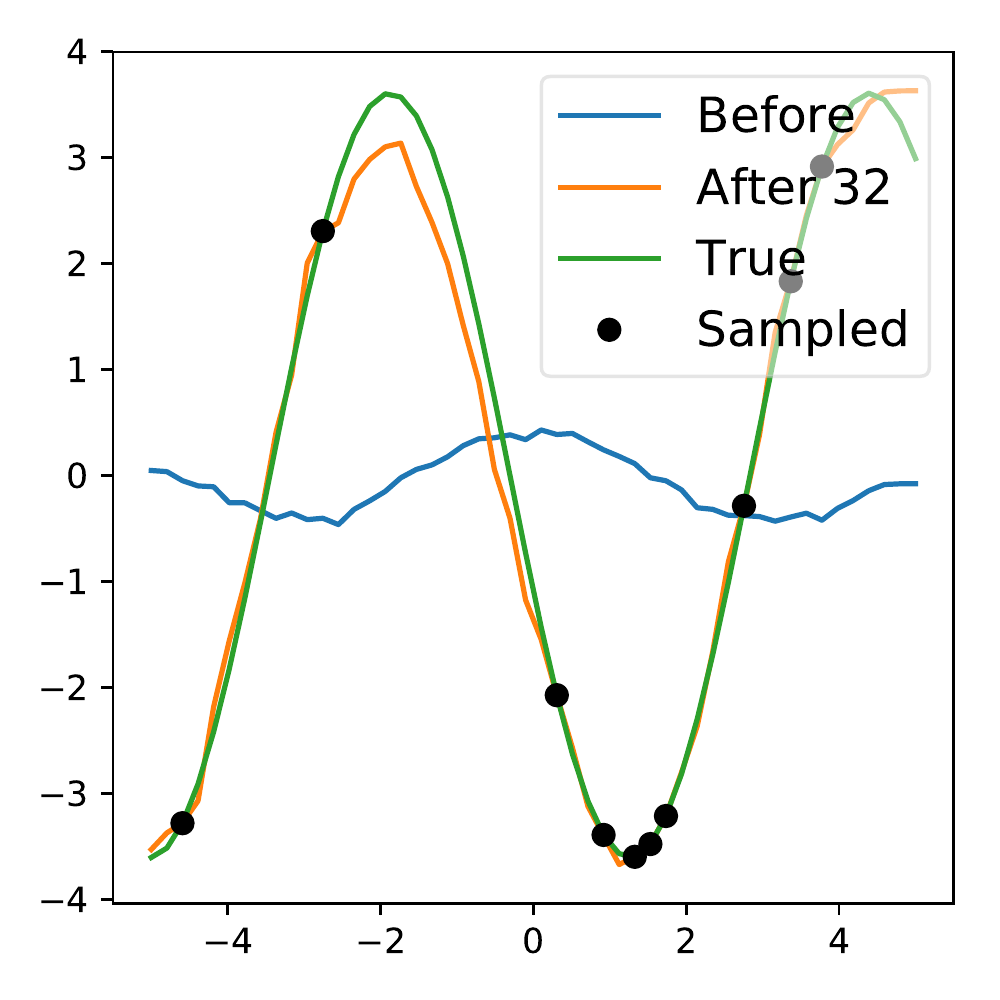}
\caption{After Reptile training}
\end{subfigure}
\caption{Demonstration of MAML and Reptile on a toy few-shot regression problem, where we train on 10 sampled points of a sine wave, performing 32 gradient steps on an MLP with layers $1 \rightarrow 64 \rightarrow 64 \rightarrow 1$.}
\label{fig:fastlearnsine}
\end{figure}

\section{Analysis} \label{analysis}

In this section, we provide two alternative explanations of why Reptile works.

\subsection{Leading Order Expansion of the Update}\label{sec:updateexp}
\newcommand{\og}{\overline{g}}
\newcommand{\oh}{\overline{H}}
\newcommand{\AG}{\mathrm{AvgGrad}}
\newcommand{\AGI}{\mathrm{AvgGradInner}}

Here, we will use a Taylor series expansion to approximate the update performed by Reptile and MAML.
We will show that both algorithms contain the same leading-order terms: the first term minimizes the expected loss (joint training), the second and more interesting term maximizes within-task generalization. Specifically, it maximizes the inner product between the gradients on different minibatches from the same task. If gradients from different batches have positive inner product, then taking a gradient step on one batch improves performance on the other batch.

Unlike in the discussion and analysis of MAML, we won't consider a training set and test set from each task; instead, we'll just assume that each task gives us a sequence of $k$ loss functions $L_1, L_2, \dots, L_{k}$; for example, classification loss on different minibatches.
We will use the following definitions:
\begin{alignat}{2}
    g_i &= L'_i(\phi_i) &&\quad\text{(gradient obtained during SGD)}\\
    \phi_{i+1} &= \phi_i - \alpha g_i &&\quad\text{(sequence of parameter vectors)}\\
    \og_i &= L'_i(\phi_1) &&\quad\text{(gradient at initial point)}\\
    \oh_i &= L''_i(\phi_1) &&\quad\text{(Hessian at initial point)}
\end{alignat}
For each of these definitions, $i \in [1, k]$.

First, let's calculate the SGD gradients to $O(\alpha^2)$ as follows.
\begin{align}
    g_i = L'_i(\phi_i) &= L'_i(\phi_1) + L''_i(\phi_1) (\phi_i - \phi_1) + \underbrace{O(\norm{\phi_i - \phi_1}^2)}_{=O(\alpha^2)} \quad\text{(Taylor's theorem)}\\
    &= \og_i + \oh_i (\phi_i - \phi_1) + O(\alpha^2) \quad\text{(using definition of $\og_i, \oh_i$)}\\
    &= \og_i - \alpha \oh_i \sum_{j=1}^{i-1} g_j + O(\alpha^2)  \quad\text{(using $\phi_i - \phi_1 = -\alpha \sum_{j=1}^{i-1} g_j$)} \\
    &= \og_i - \alpha \oh_i \sum_{j=1}^{i-1} \og_j + O(\alpha^2) \quad\text{(using $g_j = \og_j + O(\alpha)$)} \label{gradfinal}
\end{align}

Next, we will approximate the MAML gradient.
Define $U_i$ as the operator that updates the parameter vector on minibatch $i$: $U_i(\phi) = \phi - \alpha L_i'(\phi)$.
\begin{align}
    \gmaml 
    &= \dbyd{\phi_1} L_{k}(\phi_{k})\\ 
    &= \dbyd{\phi_1} L_{k}(U_{k-1}(U_{k-2}(\dots (U_1 (\phi_1)))))\\
    &=  U'_1(\phi_1)\cdots U'_{k-1}(\phi_{k-1})  L'_k(\phi_k) \qquad\text{(repeatedly applying the chain rule)} \\
    &=  \lrparen*{I - \alpha L''_1(\phi_1)}\cdots  \lrparen*{I - \alpha L''_{k-1}(\phi_{k-1})} L'_k(\phi_k) \qquad\text{(using $U'_i(\phi) = I - \alpha L''_i(\phi)$)}\\
    &=  \lrparen*{\prod_{j=1}^{k-1}(I - \alpha L''_j(\phi_j))} g_{k} \qquad\text{(product notation, definition of $g_k$)}
\end{align}
Next, let's expand to leading order
\begin{align}
    \gmaml 
    &=  \lrparen*{\prod_{j=1}^{k-1} (I - \alpha \oh_j)} \lrparen*{\og_{k} - \alpha \oh_{k} \sum_{j=1}^{k-1} \og_j} + O(\alpha^2)\\
    &\qquad\text{(replacing $L''_j(\phi_j)$ with $\oh_j$, and replacing $g_{k}$ using \Cref{gradfinal})} \nonumber\\
    &=  \lrparen*{I - \alpha \sum_{j=1}^{k-1} \oh_j} \lrparen*{\og_{k} - \alpha \oh_{k} \sum_{j=1}^{k-1} \og_j} + O(\alpha^2)\\
    &=  \og_{k} - \alpha \sum_{j=1}^{k-1} \oh_j \og_{k} -\alpha \oh_{k}  \sum_{j=1}^{k-1} \og_j+ O(\alpha^2)
\end{align}

For simplicity of exposition, let's consider the $k=2$ case, and later we'll provide the general formulas.
\begin{alignat}{2}
    \gmaml & &&= \og_2 - \alpha \oh_2 \og_1 - \alpha \oh_1 \og_2  + O(\alpha^2) \\
    \gfom &= g_2 &&= \og_2 - \alpha \oh_2 \og_1  + O(\alpha^2) \\
    \grep &= g_1 + g_2 &&= \og_1 + \og_2 - \alpha \oh_2 \og_1 + O(\alpha^2)
\end{alignat}
As we will show in the next paragraph, the terms like $\oh_2 \og_1$ serve to maximize the inner products between the gradients computed on different minibatches, while lone  gradient terms like $\og_1$ take us to the minimum of the joint training problem.

When we take the expectation of $\gfom$, $\grep$, and $\gmaml$ under minibatch sampling, we are left with only two kinds of terms which we will call $\AG$ and $\AGI$.
In the equations below $\Eb{\tau, 1, 2}{\dots}$ means that we are taking the expectation over the task $\tau$ and the two minibatches defining $L_1$ and $L_2$, respectively.
\begin{itemize}
  \item $\AG$ is defined as gradient of expected loss.
\begin{align}
\AG = \Eb{\tau, 1}{\og_1}
\end{align}
$(-\AG)$  is the direction that brings $\phi$ towards the minimum of the ``joint training'' problem; the expected loss over tasks.
\item The more interesting term is $\AGI$, defined as follows:
\begin{alignat}{2}
\AGI 
  &= \Eb{\tau, 1, 2}{\oh_2 \og_1} \\
  &= \Eb{\tau, 1, 2}{\oh_1 \og_2} &&\quad\text{(interchanging indices $1,2$)} \\
  &= \half\Eb{\tau, 1, 2}{\oh_2 \og_1 + \oh_1 \og_2} &&\quad\text{(averaging last two equations)}\\
  &=\half \Eb{\tau, 1, 2}{\dbyd{\phi_1}(\og_1 \cdot \og_2)}    
\end{alignat}
Thus, $(-\AGI)$ is the direction that increases the inner product between gradients of different minibatches for a given task, improving generalization.
\end{itemize}

Recalling our gradient expressions, we get the following expressions for the meta-gradients, for SGD with $k=2$:
\begin{align}
    \Ea{\gmaml} &= (1)\AG - (2 \alpha) \AGI + O(\alpha^2) \\
    \Ea{\gfom} &= (1)\AG - (\alpha) \AGI  + O(\alpha^2) \label{eq:fom} \\
    \Ea{\grep} &= (2) \AG - (\alpha)  \AGI + O(\alpha^2) \label{eq:rep}
\end{align}
In practice, all three gradient expressions first bring us towards the minimum of the expected loss over tasks, then the higher-order $\AGI$ term enables fast learning by maximizing the inner product between gradients within a given task.

Finally, we can extend these calculations to the general $k \ge 2$ case:
\begin{align}
    \gmaml &= \og_{k} - \alpha \oh_k \sum_{j=1}^{k-1} \og_j - \alpha \sum_{j=1}^{k-1} \oh_{j} \og_{k} + O(\alpha^2)\\
    \Ea{\gmaml} &= (1)\AG - (2(k-1)\alpha) \AGI\\
    \gfom &= g_{k}  = \og_{k} - \alpha \oh_k\sum_{j=1}^{k-1}  \og_j + O(\alpha^2)\\
    \Ea{\gfom} &= (1)\AG - ((k-1)\alpha) \AGI\\
    \grep &= -(\phi_{k+1} - \phi_1)/\alpha =  \sum_{i=1}^{k} g_i = 
    \sum_{i=1}^{k} \og_i - \alpha \sum_{i=1}^{k}\sum_{j=1}^{i-1} \oh_i \og_j + O(\alpha^2)\\
    \Ea{\grep} &= (k)  \AG - \lrparen*{\half k(k-1)\alpha} \AGI
\end{align}
As in the $k=2$, the ratio of coefficients of the $\AGI$ term and the $\AG$ term goes $\text{MAML} > \text{FOMAML} > \text{Reptile}$.
However, in all cases, this ratio increases linearly with both the stepsize $\alpha$ and the number of iterations $k$. Note that the Taylor series approximation only holds for small $\alpha k$.

\subsection{Finding a Point Near All Solution Manifolds}

Here, we argue that Reptile converges towards a solution $\phi$ that is close (in Euclidean distance) to each task $\tau$'s manifold of optimal solutions.
This is a informal argument and should be taken much less seriously than the preceding Taylor series analysis.

Let $\phi$ denote the network initialization, and let $\cW_{\tau}$ denote the set of optimal parameters for task $\tau$. We want to find $\phi$ such that the distance $D(\phi, \cW_{\tau})$ is small for all tasks. 
\begin{align}
\operatorname*{minimize}_{\phi} \Eb{\tau}{\half D(\phi, \cW_{\tau})^2}
\end{align}
We will show that Reptile corresponds to performing SGD on that objective. 

Given a non-pathological set $S \subset \Real^d$, then for almost all points $\phi \in \Real^d$ the gradient of the squared distance $D(\phi, S)^2$ is $2(\phi - P_S(\phi))$, where $P_S(\phi)$ is the projection (closest point) of $\phi$ onto $S$.
Thus,
\begin{align}
\gradph \Eb{\tau}{\half D(\phi, \cW_{\tau})^2}
&= \Eb{\tau}{\half \gradph D(\phi, \cW_{\tau})^2}\\
&= \Eb{\tau}{\phi - P_{\cW_{\tau}}(\phi)}, \text{where}\ P_{\cW_{\tau}}(\phi) = \argmin_{p \in \cW_{\tau}} D(p, \phi)
\end{align}
Each iteration of Reptile corresponds to sampling a task $\tau$ and performing a stochastic gradient update
\begin{align}
\phi &\leftarrow \phi - \epsilon \nabla_{\phi} \tfrac{1}{2}D(\phi, \cW_{\tau})^2 \\
&\quad= \phi - \epsilon(\phi - P_{\cW_{\tau}}(\phi) )\\
&\quad= (1-\epsilon)\phi + \epsilon P_{\cW_{\tau}}(\phi).
\end{align}
In practice, we can't exactly compute $P_{\cW_{\tau}}(\phi)$, which is defined as a minimizer of $\lt$.
However, we can partially minimize this loss using gradient descent. Hence, in Reptile we replace $W^{\ast}_{\tau}(\phi)$ by the result of running $k$ steps of gradient descent on $L_{\tau}$ starting with initialization $\phi$.

\begin{figure}
\centering
\includegraphics[width=.4\textwidth]{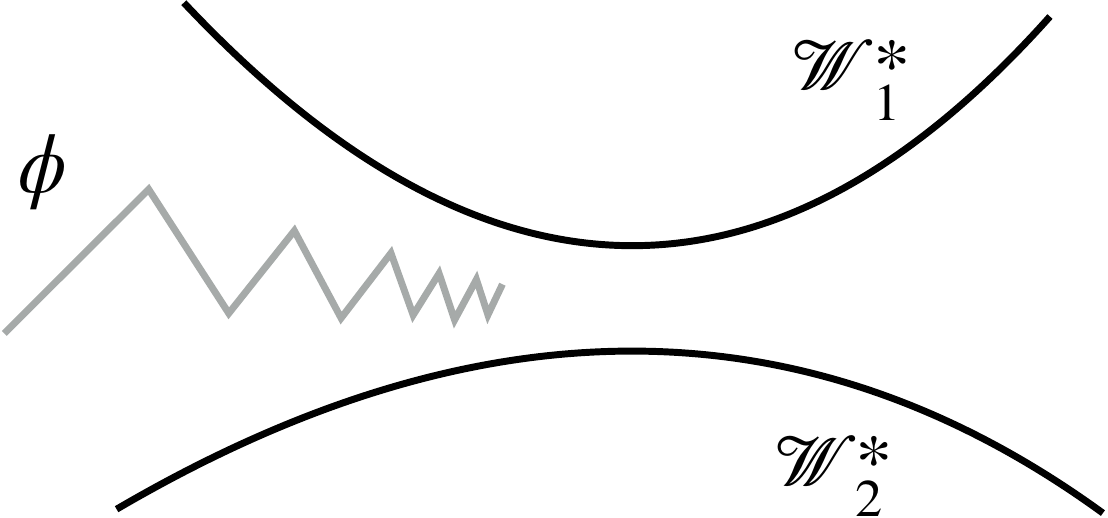}
\caption{The above illustration shows the sequence of iterates obtained by moving alternately towards two optimal solution manifolds $\cW_1$ and $\cW_2$ and converging to the point that minimizes the average squared distance. One might object to this picture on the grounds that we converge to the same point regardless of whether we perform one step or multiple steps of gradient descent. That statement is true, however, note that minimizing the expected distance objective $\Eb{\tau}{D(\phi, \cW_{\tau})}$ is different than minimizing the expected loss objective $\Eb{\tau}{\lt(f_{\phi})}$. In particular, there is a high-dimensional manifold of minimizers of the expected loss $\lt$ (e.g., in the sine wave case, many neural network parameters give the zero function $f(\phi)=0$), but the minimizer of the expected distance objective is typically a single point.}
\end{figure}

\section{Experiments}

\subsection{Few-Shot Classification}
\label{sec:fewshot}

We evaluate our method on two popular few-shot classification tasks: Omniglot \cite{lake2011one} and Mini-ImageNet \cite{vinyals2016matching}. These datasets make it easy to compare our method to other few-shot learning approaches like MAML.

In few-shot classification tasks, we have a meta-dataset $D$ containing many classes $C$, where each class is itself a set of example instances $\{c_1, c_2, ..., c_n\}$. If we are doing $K$-shot, $N$-way classification, then we sample tasks by selecting $N$ classes from $C$ and then selecting $K+1$ examples for each class. We split these examples into a training set and a test set, where the test set contains a single example for each class. The model gets to see the entire training set, and then it must classify a randomly chosen sample from the test set. For example, if you trained a model for 5-shot, 5-way classification, then you would show it 25 examples (5 per class) and ask it to classify a 26\textsuperscript{th} example.

In addition to the above setup, we also experimented with the \textit{transductive} setting, where the model classifies the entire test set at once. In our transductive experiments, information was shared between the test samples via batch normalization \cite{ioffe2015batch}. In our non-transductive experiments, batch normalization statistics were computed using all of the training samples and a single test sample. We note that Finn et al. \cite{finn2017model} use transduction for evaluating MAML.

For our experiments, we used the same CNN architectures and data preprocessing as Finn et al. \cite{finn2017model}. We used the Adam optimizer \cite{kingma2015adam} in the inner loop, and vanilla SGD in the outer loop, throughout our experiments. For Adam we set $\beta_1 = 0$ because we found that momentum reduced performance across the board.\footnote{This finding also matches our analysis from \Cref{sec:updateexp}, which suggests that Reptile works because sequential steps come from different mini-batches. With momentum, a mini-batch has influence over the next few steps, reducing this effect.} During training, we never reset or interpolated Adam's rolling moment data; instead, we let it update automatically at every inner-loop training step. However, we did backup and reset the Adam statistics when evaluating on the test set to avoid information leakage.

The results on Omniglot and Mini-ImageNet are shown in \Cref{tab:mini-img,tab:omni}. While MAML, FOMAML, and Reptile have very similar performance on all of these tasks, Reptile does slightly better than the alternatives on Mini-ImageNet and slightly worse on Omniglot. It also seems that transduction gives a performance boost in all cases, suggesting that further research should pay close attention to its use of batch normalization during testing.

\begin{table}[H]
    \begin{center}
        \begin{tabular}{|l|c|c|c|c|}
            \hline
            Algorithm & 1-shot 5-way & 5-shot 5-way \\
            \hline
            MAML + Transduction & $48.70 \pm 1.84\%$ & $63.11 \pm 0.92\%$ \\
            \hline
            \first-order MAML + Transduction & $48.07 \pm 1.75\%$ & $63.15 \pm 0.91\%$ \\
            \hline
            Reptile & $47.07 \pm 0.26\%$ & $62.74 \pm 0.37\%$ \\
            \hline
            Reptile + Transduction & $49.97 \pm 0.32\%$ & $65.99 \pm 0.58\%$ \\
            \hline
        \end{tabular}
    \end{center}
    \caption {Results on Mini-ImageNet. Both MAML and \first-order MAML results are from \cite{finn2017model}.} \label{tab:mini-img}
\end{table}

\begin{table}[H]
    \begin{center}
        \begin{tabular}{|l|c|c|c|c|}
            \hline
            Algorithm & 1-shot 5-way & 5-shot 5-way & 1-shot 20-way & 5-shot 20-way \\
            \hline
            MAML + Transduction & $98.7 \pm 0.4\%$ & $99.9 \pm 0.1\%$ & $95.8 \pm 0.3\%$ & $98.9 \pm 0.2\%$ \\
            \hline
            \first-order MAML + Transduction & $98.3 \pm 0.5\%$ & $99.2 \pm 0.2\%$ & $89.4 \pm 0.5\%$ & $97.9 \pm 0.1\%$ \\
            \hline
            Reptile & $95.39 \pm 0.09\%$ & $98.90 \pm 0.10\%$ & $88.14 \pm 0.15\%$ & $96.65 \pm 0.33\%$ \\
            \hline
            Reptile + Transduction & $97.68 \pm 0.04\%$ & $99.48 \pm 0.06\%$ & $89.43 \pm 0.14\%$ & $97.12 \pm 0.32\%$ \\
            \hline
        \end{tabular}
    \end{center}
    \caption {Results on Omniglot. MAML results are from \cite{finn2017model}. \first-order MAML results were generated by the code for \cite{finn2017model} with the same hyper-parameters as MAML.} \label{tab:omni}
\end{table}

\subsection{Comparing Different Inner-Loop Gradient Combinations} \label{sec:comparinginner}

For this experiment, we used four non-overlapping mini-batches in each inner-loop, yielding gradients $g_1$, $g_2$, $g_3$, and $g_4$. We then compared learning performance when using different linear combinations of the $g_i$'s for the outer loop update. Note that two-step Reptile corresponds to $g_1 + g_2$, and two-step FOMAML corresponds to $g_2$.

To make it easier to get an apples-to-apples comparison between different linear combinations, we simplified our experimental setup in several ways. First, we used vanilla SGD in the inner- and outer-loops. Second, we did not use meta-batches. Third, we restricted our experiments to 5-shot, 5-way Omniglot. With these simplifications, we did not have to worry as much about the effects of hyper-parameters or optimizers.

\begin{figure}[h]
    \centering
    \includegraphics[width=12cm]{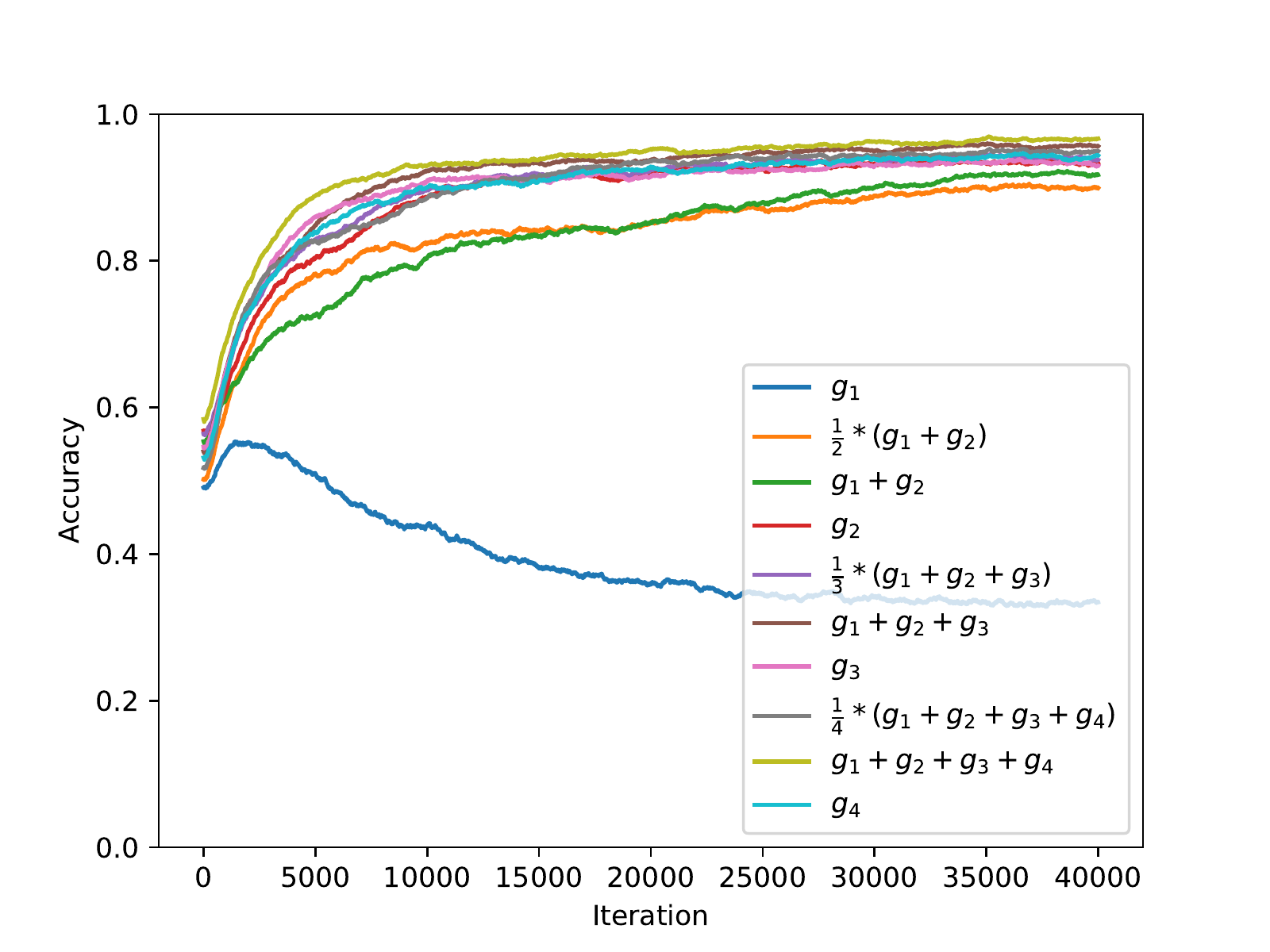}
    \caption{Different inner-loop gradient combinations on 5-shot 5-way Omniglot.}
    \label{fig:gradcomp}
\end{figure}

Figure \ref{fig:gradcomp} shows the learning curves for various inner-loop gradient combinations. For gradient combinations with more than one term, we ran both a sum and an average of the inner gradients to correct for the effective step size increase.

As expected, using only the first gradient $g_1$ is quite ineffective, since it amounts to optimizing the expected loss over all tasks. Surprisingly, two-step Reptile is noticeably worse than two-step FOMAML, which might be explained by the fact that two-step Reptile puts less weight on $\AGI$ relative to $\AG$ (\Cref{eq:fom,eq:rep}). Most importantly, though, all the methods improve as the number of mini-batches increases. This improvement is more significant when using a sum of all gradients (Reptile) rather than using just the final gradient (FOMAML). This also suggests that Reptile can benefit from taking many inner loop steps, which is consistent with the optimal hyper-parameters found for Section~\ref{sec:fewshot}.

\subsection{Overlap Between Inner-Loop Mini-Batches} \label{sec:effectsbatching}

Both Reptile and FOMAML use stochastic optimization in their inner-loops. Small changes to this optimization procedure can lead to large changes in final performance. 
This section explores the sensitivity of Reptile and FOMAML to the inner loop hyperparameters, and also shows that FOMAML's performance significantly drops if mini-batches are selected the wrong way.

The experiments in this section look at the difference between \textit{shared-tail FOMAML}, where the final inner-loop mini-batch comes from the same set of data as the earlier inner-loop batches, to \textit{separate-tail FOMAML}, where the final mini-batch comes from a disjoint set of data. 
Viewing FOMAML as an approximation to MAML, \textit{separate-tail FOMAML} can be seen as the more correct approach (and was used by Finn et al. \cite{finn2017model}), since the training-time optimization resembles the test-time optimization (where the test set doesn't overlap with the training set).
Indeed, we find that separate-tail FOMAML is significantly better than shared-tail FOMAML.
As we will show, shared-tail FOMAML degrades in performance when the data used to compute the meta-gradient ($\gfom=g_k$) overlaps significantly with the earlier batches;
however, Reptile and separate-tail MAML maintain performance and are not very sensitive to the inner-loop hyperparameters.
\begin{figure}
    \centering
    \begin{subfigure}{.3\textwidth}
        \centering
        \includegraphics[width=\linewidth]{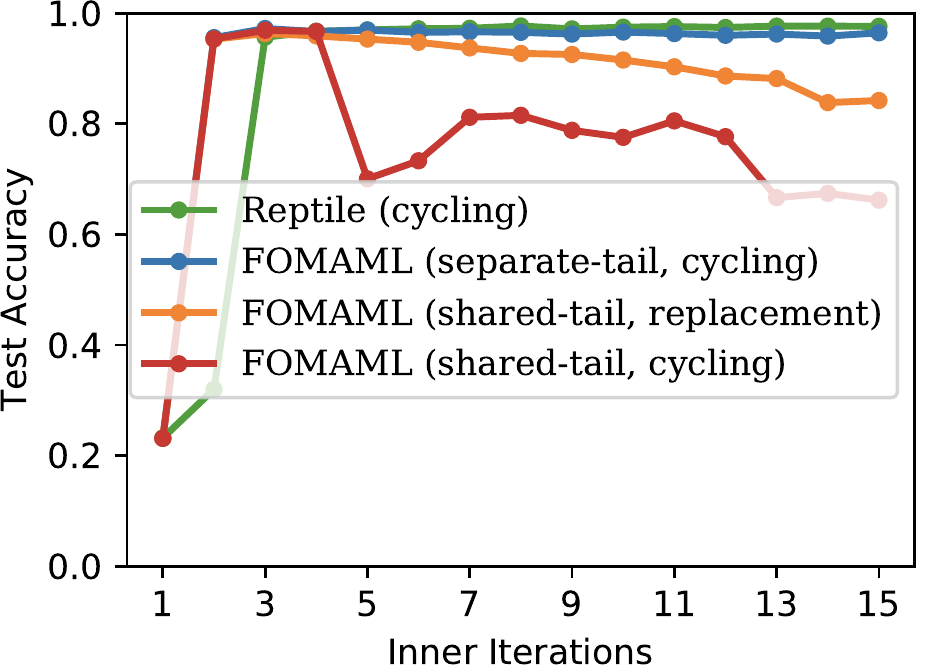}
        \caption{Final test performance vs. number of inner-loop iterations.}
        \label{fig:hp_niters}
    \end{subfigure}
    \hfill
    \begin{subfigure}{.3\textwidth}
        \centering
        \includegraphics[width=\linewidth]{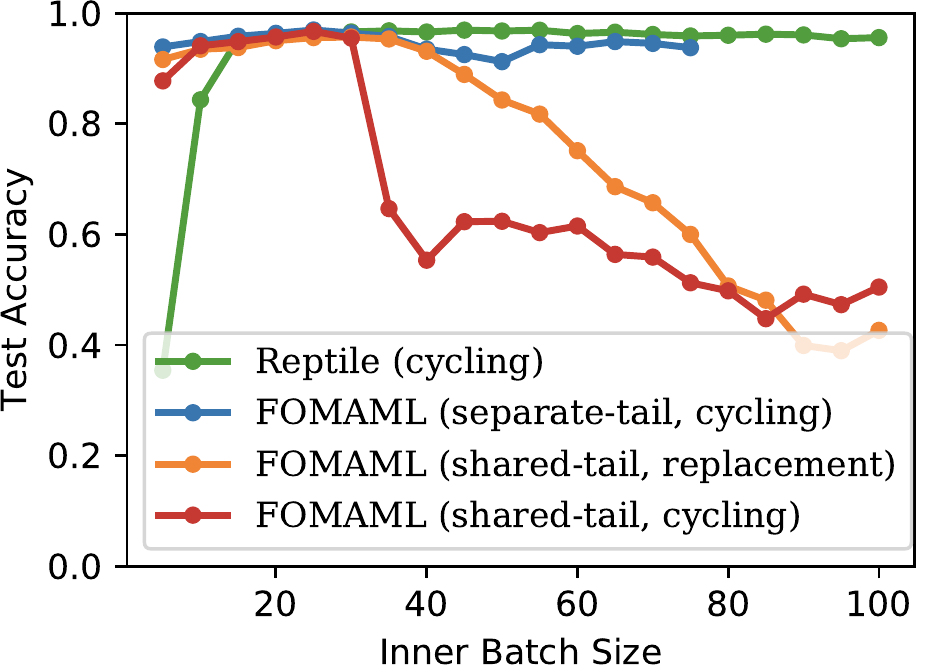}
        \caption{Final test performance vs. inner-loop batch size.}
        \label{fig:hp_batch}
    \end{subfigure}    
    \hfill
    \begin{subfigure}{.3\textwidth}
        \centering
        \includegraphics[width=\linewidth]{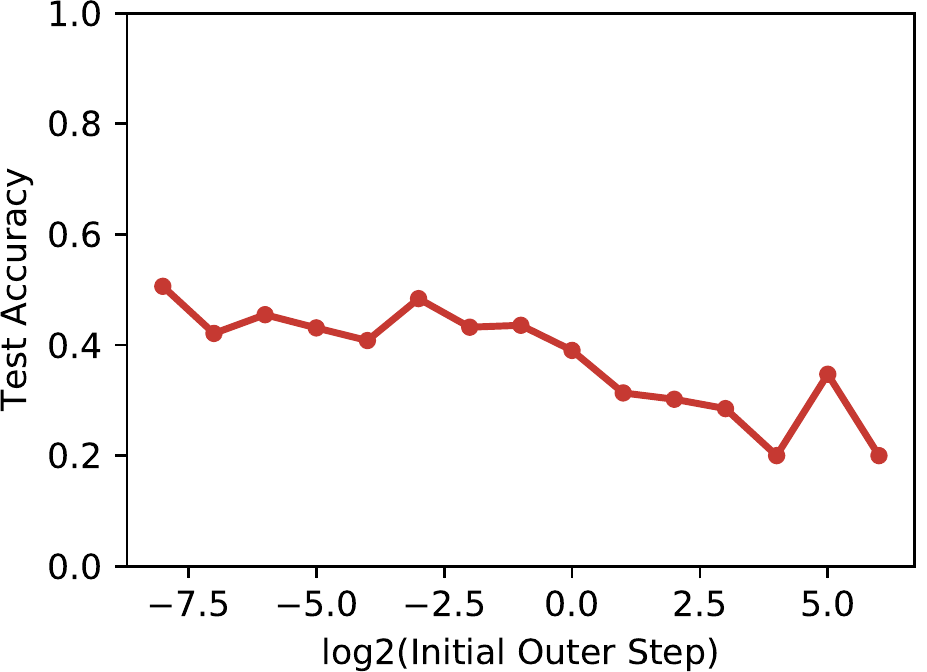}
        \caption{Final test performance vs. outer-loop step size for shared-tail FOMAML with batch size 100 (full batches).}
        \label{fig:hp_stepsize}
    \end{subfigure}
    \caption{The results of hyper-parameter sweeps on 5-shot 5-way Omniglot.}
\end{figure}

\Cref{fig:hp_niters} shows that when minibatches are selected by cycling through the training data (shared-tail, cycle), shared-tail FOMAML performs well up to four inner-loop iterations, but drops in performance starting at five iterations, where the final minibatch (used to compute $\gfom = g_k$) overlaps with the earlier ones.
When we use random sampling instead (shared-tail, replacement), shared-tail FOMAML degrades more gradually. We hypothesize that this is because some samples still appear in the final batch that were not in the previous batches. The effect is stochastic, so it makes sense that the curve is smoother.

\Cref{fig:hp_batch} shows a similar phenomenon, but here we fixed the inner-loop to four iterations and instead varied the batch size. For batch sizes greater than 25, the final inner-loop batch for shared-tail FOMAML necessarily contains samples from the previous batches. Similar to \Cref{fig:hp_niters}, here we observe that shared-tail FOMAML with random sampling degrades more gradually than shared-tail FOMAML with cycling.

In both of these parameter sweeps, separate-tail FOMAML and Reptile do not degrade in performance as the number of inner-loop iterations or batch size changes. 

There are several possible explanations for above findings. For example, one might hypothesize that shared-tail FOMAML is only worse in these experiments because its effective step size is much lower than that of separate-tail FOMAML. However, \Cref{fig:hp_stepsize} suggests that this is not the case: performance was equally poor for every choice of step size in a thorough sweep. A different hypothesis is that shared-tail FOMAML performs poorly because, after a few inner-loop steps on a sample, the gradient of the loss for that sample does not contain very much useful information about the sample. In other words, the first few SGD steps might bring the model close to a local optimum, and then further SGD steps might simply bounce around this local optimum.

\section{Discussion}

Meta-learning algorithms that perform gradient descent at test time are appealing because of their simplicity and generalization properties \cite{finn2017meta}.
The effectiveness of fine-tuning (e.g. from models trained on ImageNet \cite{deng2009imagenet}) gives us additional faith in these approaches.
This paper proposed a new algorithm called Reptile, whose training process is only subtlely different from joint training and only uses first-order gradient information (like first-order MAML).

We gave two theoretical explanations for why Reptile works.
First, by approximating the update with a Taylor series, we showed that SGD \textit{automatically} gives us the same kind of second-order term that MAML computes.
This term adjusts the initial weights to maximize the dot product between the gradients of different minibatches on the same task---i.e., it encourages the gradients to generalize between minibatches of the same task.
We also provided a second informal argument, which is that Reptile finds a point that is close (in Euclidean distance) to all of the optimal solution manifolds of the training tasks.

While this paper studies the meta-learning setting, the Taylor series analysis in \Cref{sec:updateexp} may have some bearing on stochastic gradient descent in general. 
It suggests that when doing stochastic gradient descent, we are automatically performing a MAML-like update that maximizes the generalization between different minibatches.
This observation partly explains why fine tuning (e.g., from ImageNet to a smaller dataset \cite{zhang2014part}) works well.
This hypothesis would suggest that \textit{joint training plus fine tuning} will continue to be a strong baseline for meta-learning in various machine learning problems.

\section{Future Work}

We see several promising directions for future work:
\begin{itemize}
  \item Understanding to what extent SGD automatically optimizes for generalization, and whether this effect can be amplified in the non-meta-learning setting.
  \item Applying Reptile in the reinforcement learning setting. So far, we have obtained negative results, since joint training is a strong baseline, so some modifications to Reptile might be necessary.
  \item Exploring whether Reptile's few-shot learning performance can be improved by deeper architectures for the classifier.
  \item Exploring whether regularization can improve few-shot learning performance, as currently there is a large gap between training and testing error.
  \item Evaluating Reptile on the task of few-shot density modeling \cite{reed2017few}.
\end{itemize}


{\bibliographystyle{plain}
\small
\bibliography{reptile}
}

\appendix

\section{Hyper-parameters}

For all experiments, we linearly annealed the outer step size to 0. We ran each experiment with three different random seeds, and computed the confidence intervals using the standard deviation across the runs.

Initially, we tried optimizing the Reptile hyper-parameters using CMA-ES \cite{hansen2006cma}. However, we found that most hyper-parameters had little effect on the resulting performance. After seeing this result, we simplified all of the hyper-parameters and shared hyper-parameters between experiments when it made sense.

\begin{table}[H]
    \caption {Reptile hyper-parameters for the Omniglot comparison between all algorithms.} \label{tab:title}
    \begin{center}
        \begin{tabular}{|c|c|c|c|c|}
            \hline
            Parameter & 5-way & 20-way \\
            \hline
            Adam learning rate & 0.001 & 0.0005 \\
            Inner batch size & 10 & 20 \\
            Inner iterations & 5 & 10 \\
            Training shots & 10 & 10 \\
            \hline
            Outer step size & 1.0 & 1.0 \\
            Outer iterations & 100K & 200K \\
            Meta-batch size & 5 & 5 \\
            \hline
            Eval. inner iterations & 50 & 50 \\
            Eval. inner batch & 5 & 10 \\
            \hline
        \end{tabular}
    \end{center}
\end{table}

\begin{table}[H]
    \caption {Reptile hyper-parameters for the Mini-ImageNet comparison between all algorithms.} \label{tab:title}
    \begin{center}
        \begin{tabular}{|c|c|c|}
            \hline
            Parameter & 1-shot & 5-shot \\
            \hline
            Adam learning rate & $0.001$ & $0.001$ \\
            Inner batch size & 10 & 10 \\
            Inner iterations & 8 & 8 \\
            Training shots & 15 & 15 \\
            \hline
            Outer step size & 1.0 & 1.0 \\
            Outer iterations & 100K & 100K \\
            Meta-batch size & 5 & 5 \\
            \hline
            Eval. inner batch size & 5 & 15 \\
            Eval. inner iterations & 50 & 50 \\
            \hline
        \end{tabular}
    \end{center}
\end{table}

\begin{table}[H]
    \caption {Hyper-parameters for \Cref{sec:comparinginner}. All outer step sizes were linearly annealed to zero during training.} \label{tab:title}
    \begin{center}
        \begin{tabular}{|c|c|}
            \hline
            Parameter & Value \\
            \hline
            Inner learning rate & $3 \times 10^{-3}$ \\
            Inner batch size & 25 \\
            \hline
            Outer step size & 0.25 \\
            Outer iterations & 40K \\
            \hline
            Eval. inner batch size & 25 \\
            Eval. inner iterations & 5 \\
            \hline
        \end{tabular}
    \end{center}
\end{table}

\begin{table}[H]
    \caption {Hyper-parameters \Cref{sec:effectsbatching}. All outer step sizes were linearly annealed to zero during training.} \label{tab:title}
    \begin{center}
        \begin{tabular}{|c|c|c|c|}
            \hline
            Parameter & \Cref{fig:hp_batch} & \Cref{fig:hp_niters} & \Cref{fig:hp_stepsize} \\
            \hline
            Inner learning rate & $3 \times 10^{-3}$ & $3 \times 10^{-3}$ & $3 \times 10^{-3}$ \\
            Inner batch size & - & 25 & 100 \\
            Inner iterations & 4 & - & 4 \\
            \hline
            Outer step size & 1.0 & 1.0 & - \\
            Outer iterations & 40K & 40K & 40K \\
            \hline
            Eval. inner batch size & 25 & 25 & 25 \\
            Eval. inner iterations & 5 & 5 & 5 \\
            \hline
        \end{tabular}
    \end{center}
\end{table}

\end{document}

%% file: rl_math.tex
\newcommand{\Real}{\mathbb{R}}
\DeclareMathOperator*{\argmin}{arg\,min}

\DeclareMathOperator*{\minimize}{minimize}

\newcommand{\half}{\frac{1}{2}}

\newcommand{\grad}{\nabla}

\newcommand{\E}{\mathbb{E}}

\newcommand{\Ea}[1]{\E\left[#1\right]}
\newcommand{\Eb}[2]{\E_{#1}\left[#2\right]}

\DeclarePairedDelimiter{\norm}{\|}{\|}

\DeclarePairedDelimiter{\lrparen}{(}{)}



%% file: exercises_def.tex
\newcommand{\beq}{\begin{equation}}
\newcommand{\eeq}{\end{equation}}
\newcommand{\bea}{\begin{eqnarray}}
\newcommand{\eea}{\end{eqnarray}}
\newcommand{\beas}{\begin{eqnarray*}}
\newcommand{\eeas}{\end{eqnarray*}}
\newcommand{\ea}{\end{array}}
\newcommand{\bit}{\begin{itemize}}
\newcommand{\eit}{\end{itemize}}
\newcommand{\ben}{\begin{enumerate}}
\newcommand{\een}{\end{enumerate}}
\newcommand{\bde}{\begin{description}}
\newcommand{\ede}{\end{description}}
\newcommand{\bsp}{\begin{split}}
\newcommand{\esp}{\end{split}}

\newcommand{\monthyear}{%
  \ifcase\month\or January\or February\or March\or April\or May\or June\or
  July\or August\or September\or October\or November\or
  December\fi\space\number\year
}

